\documentclass[sigconf,nonacm]{acmart}

\setcopyright{none}
\acmYear{}
\acmDOI{}
\acmISBN{}
\acmConference{}{}{}
\acmBooktitle{}
\acmPrice{}

\usepackage{booktabs}
\usepackage{graphicx}
\usepackage{array}
\usepackage{amsmath}
\usepackage{multirow}
\usepackage{longtable}
\usepackage{hyperref}

\AtBeginDocument{\providecommand\BibTeX{{B\kern-.05em{\sc i\kern-.025em b}\kern-.08em\TeX}}}

\begin{document}

\title{Explainable Detection of AI-Generated Images with Artifact Localization Using Faster-Than-Lies and Vision--Language Models for Edge Devices}

\author{Aryan Mathur}
\affiliation{
  \institution{Indian Institute of Technology Palakkad, India}
  \country{}
}
\email{aryannmathur@gmail.com}

\author{Asaduddin Ahmed}
\affiliation{
  \institution{Indian Institute of Technology Palakkad, India}
  \country{}
}
\email{ahmedasad72425@gmail.com}

\author{Pushti Amit Vasoya}
\affiliation{
  \institution{Indian Institute of Technology Palakkad, India}
  \country{}
}
\email{vasoyapushti@gmail.com}

\author{Simeon Kandan Sonar}
\affiliation{
  \institution{Indian Institute of Technology Palakkad, India}
  \country{}
}
\email{simeon13072005@gmail.com}

\author{Yasir Z}

\affiliation{
  \institution{Indian Institute of Technology Palakkad, India}
  \country{}
}
\email{mazakhus2@gmail.com}

\author{Madesh Kuppusamy}
\affiliation{
  \institution{Indian Institute of Technology Palakkad, India}
  \country{}
}
\email{madesh6805@gmail.com}

\begin{abstract}
The increasing realism of AI-generated imagery poses challenges for verifying visual authenticity. We present an explainable image authenticity detection system that combines a lightweight convolutional classifier (“Faster-Than-Lies”) with a Vision–Language Model (Qwen2-VL-7B) to classify, localize, and explain artifacts in 32×32 images. Our model achieves 96.5\% accuracy on the extended CiFAKE dataset augmented with adversarial perturbations and maintains an inference time of 175 ms on 8-core CPUs, enabling deployment on local or edge devices. Using autoencoder-based reconstruction error maps, we generate artifact localization heatmaps, which enhance interpretability for both humans and the VLM. We further categorize 70 visual artifact types into eight semantic groups and demonstrate explainable text generation for each detected anomaly. This work highlights the feasibility of combining visual and linguistic reasoning for interpretable authenticity detection in low-resolution imagery and outlines potential cross-domain applications in forensics, industrial inspection, and social media moderation.
\end{abstract}

\maketitle
\settopmatter{printacmref=false}



\section*{Introduction}
In an age where AI can generate hyper-realistic images, distinguishing between what’s real and what’s artificially generated has become more challenging than ever. What if we could not only detect these artificial creations but also understand the reasoning behind that detection? This is the heart of our project: using the power of Faster Than Lies (FTL), and Vision-Language Models (VLMs) to not only unravel the mystery of real vs. fake images but also provide clear, human-understandable explanations behind every decision.

Our approach goes beyond just labeling  32*32 images. By using FTL, we ensure that the model’s decision-making process is transparent and interpretable, shedding light on the reasons behind each classification. VLMs take this a step further, transforming technical findings into natural language, helping anyone—regardless of their AI expertise—grasp why an image is flagged as fake. With this powerful combination, our system offers more than just accuracy: it fosters trust and understanding in AI's role in discerning the authenticity of visual content in an increasingly synthetic world

\section*{Key Challenges}

Developing an efficient model for classifying and explaining 32×32 resolution images presented a range of unique challenges. Most existing research focuses on high-resolution images, meaning that adapting these methodologies to such a low resolution required innovative approaches and extensive experimentation. Below, we outline the primary obstacles encountered during the development process.

\subsection*{Model Design for Low-Resolution Images}
Classifying 32×32 pixel images was particularly challenging due to the limited amount of visual information they provided. Traditional classifiers designed for higher-resolution images, such as 256x256 and beyond, proved ineffective when applied to lower-resolution images. This necessitated the development of novel architectures and techniques. Achieving satisfactory performance required multiple rounds of experimentation and model refinement. We narrowed down to \textbf{CNNs and Vision Transformers}

\subsection*{Lightweight and Locally Deployable Model}
Another critical challenge was designing a lightweight model that could be efficiently deployed and run on local devices, as specified in the problem statement. The model needed to meet the following requirements: 
\begin{itemize}
    \item \textbf{Parameters:} approximately 23 million
    \item \textbf{Inference time:} approximately 200 ms on CPUs (excluding model loading time)
    \item \textbf{Size:} approximately 98 MB
\end{itemize}
Achieving a balance between computational efficiency and high accuracy was paramount. In particular, it was crucial that the model maintained reliability while being deployable on devices with limited resources, such as edge devices or low-power systems. This challenge also required addressing real-time inference needs without sacrificing performance.

To meet these requirements, we explored various state-of-the-art models, including:

\begin{itemize}
    \item \textbf{EfficientNet B0, B1, B2, B3:} These models are known for their efficiency in terms of parameters and computational requirements, making them strong candidates for low-resource environments.
    \item \textbf{ResNet32:} A lightweight convolutional network with fewer parameters compared to traditional deep networks, providing a good trade-off between speed and accuracy.
    \item \textbf{Vision Transformers (ViT):} Vision Transformers have shown excellent performance in vision tasks but typically require more computational resources. We explored various variants to determine if a lightweight version could fit our needs.
    \item \textbf{Top Models from Each Category:} We also investigated leading models from each domain, including lightweight convolutional neural networks (CNNs) and transformers optimized for vision tasks.
\end{itemize}

After extensive evaluation of these models, we ultimately selected \textbf{Faster than lies}, which has 29.8 million parameters and an inference time of approximately 175ms on an 8-core CPU. Despite having slightly more parameters than our initial target (23 million), it struck an optimal balance between speed and accuracy. Its performance on local devices was highly satisfactory, with inference times well below the required 200 ms, and its size (approximately 98 MB) was ideal for deployment in low-resource environments.

\subsection*{Explainability of Classifications}
Given the presence of various artifacts in the dataset, providing explainable classifications was a major priority. Achieving this required the integration of a fine-tuned Vision-Language Model (VLM). However, due to the low resolution of the images, the performance of the VLM in generating accurate and meaningful explanations initially fell short. To overcome this limitation, extensive prompt engineering and fine-tuning with auxiliary datasets were required to improve the model's ability to explain its predictions.

\subsection*{Image Quality}
Since the models had been downsampled to 32x32 it was very difficult for models to understand the features. So we tried upscaling and restoring the images using Non GAN and Non Diffusion techniques. We found SwinIR as one such approach that could restore the images to 128x128 with precision but this happened to add edge based artifacts (when tested on a localisation approach we went ahead with, will be explained later in the paper). We even tried removing these artifacts but this further added complexities in more artifact identification than they had.

\begin{figure}[htbp]
  \centering
  \includegraphics[width=0.3\textwidth]{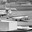}
  \caption{Image Quality since downsampled to 32x32}
  \label{fig:image_label}
\end{figure}

\subsection*{Dealing with Perturbations}

Addressing the impact of perturbations on image classification was a major challenge in our study. Initially, we focused on analyzing test images to identify various types of perturbations, using methods such as frequency analysis and Local Binary Patterns (LBP). However, we soon realized that adversarial perturbations significantly impacted the model's performance, especially when these perturbations exceeded certain thresholds.

To tackle this issue, we followed a structured process:
\begin{enumerate}
    \item First, we analyzed the perturbations in a set of test images, evaluating their intensity based on noise, compression artifacts, and blur.
    \item Next, we extended this analysis to real-world images, as well as perturbed real images, to identify perturbations that exceeded a predefined threshold.
    \item Based on this analysis, we introduced similar perturbations to our training dataset, simulating realistic conditions and challenges.
    \item Finally, we retrained our model using this augmented dataset. By exposing the model to perturbed images, we allowed it to adapt to these variations and improve its robustness to such disturbances.
\end{enumerate}

\subsection*{Choosing the Most Suitable Vision-Language Model}

Selecting the right Vision-Language Model (VLM) for the task of low-resolution image classification and artifact explanation was a critical step in our research. It involved extensive research and experimentation with multiple models, each of which was carefully evaluated against the specific requirements of the task. These requirements included the ability to handle the unique challenges posed by 32×32 resolution images and the need to generate accurate and explainable classifications of image artifacts.

We initially explored several potential models, including PixTral, Palligemma2, Llama2, OV, LLaVA7B, LLaVA13B, and Qwen2 VL 7B. Each model was tested for its performance on low-resolution images and its ability to provide meaningful explanations for classifications. The models were also assessed for computational efficiency, as running the model locally on edge devices required a balance between accuracy and resource consumption.

Among these options, Qwen2 VL 7B stood out as the most promising candidate. After conducting a series of experiments, it became clear that Qwen2 VL 7B provided the best combination of performance, accuracy, and explainability. The model demonstrated a strong ability to handle the intricacies of low-resolution images while also providing insightful explanations for the detected artifacts. Furthermore, Qwen2 VL 7B was well-optimized for deployment in a local setting, which was essential for ensuring that the model could run efficiently on resource-constrained devices.

\begin{table}[ht]
\centering
\resizebox{0.5\textwidth}{!}{%
\begin{tabular}{|>{\raggedright\arraybackslash}p{2.5cm}|>{\raggedright\arraybackslash}p{3.5cm}|>{\raggedright\arraybackslash}p{4cm}|}
\hline
\textbf{Model} & \textbf{Performance Metrics} & \textbf{Use Case Compatibility} \\ \hline
PixTral (3B) & Accuracy: 79\%, Speed: Fast & Low-resolution images and artifacts \\ \hline
Palligemma2 (4.2B) & Accuracy: 82\%, Speed: Moderate & Text-heavy use cases \\ \hline
Llama2 OV (7B) & Accuracy: 83\%, Speed: Moderate & NLP-heavy applications \\ \hline
Llava7B & Accuracy: 85\%, Speed: Moderate & Vision-Language tasks \\ \hline
Llava13B & Accuracy: 87\%, Speed: Slow & Vision-Language tasks, high accuracy \\ \hline
Qwen2 VL 7B & Accuracy: 90\%, Speed: Fast & Ideal for low-resolution image classification and explanation \\ \hline
Qwen2 VL 7B 4-bit & Accuracy: 88\%, Speed: Very Fast & Low-power devices, edge computing \\ \hline
\end{tabular}
}
\caption{Comparison of Vision-Language Models}
\end{table}

In conclusion, Qwen2 VL 7B was selected post testing process, where each potential model was evaluated based on its compatibility with the task's requirements. This model's ability to effectively handle low-resolution images and provide clear explanations made it the best choice for our system, and its performance was key to the success of our approach.

\section*{Related Work}
Prior efforts in synthetic image detection have focused on convolutional neural networks and frequency-based artifact analysis~\cite{guo2018countering, xu2024fakeshield}. However, such models often lack interpretability. Transformer-based models such as ViT~\cite{dosovitskiy2021image} and hybrid architectures improve accuracy but remain computationally expensive. Vision–Language Models (VLMs) like LLaVA~\cite{liu2023llava} and Qwen~\cite{qwen} have demonstrated strong reasoning capabilities by aligning visual and textual representations, offering promise for explainable AI. Our work integrates these paradigms by coupling a lightweight classifier (“Faster Than Lies”) with a VLM to achieve both high accuracy and human-understandable explanations, tailored for low-resolution inputs.

\section*{Methodology}
The overall objective of the problem statement being to identify if an image is AI Generated that has been generated using Stable Diffusion, PixArt, GigaGAN and Adobe Firefly with adversarial perturbations introduced to prevent detection. 
\subsubsection*{\textbf{Dataset Collection}}
We were provided with an initial dataset of CiFake (1,200,000 diffusion images). To cover other types of AI-generated images as well, we looked for images generated by specific models. We included GigaGAN Conditioned, GigaGAN T2I COCO DiffNoised, PixArt, and the ArtiFact dataset. This brought the total to 1.2 million images, with nearly equal numbers of real and fake images, i.e., around 680k fake images and 520k real images.

\subsection*{Task-1:}
\subsubsection*{Workflow}
\begin{figure}[htbp]
  \centering
  \includegraphics[scale=0.3]{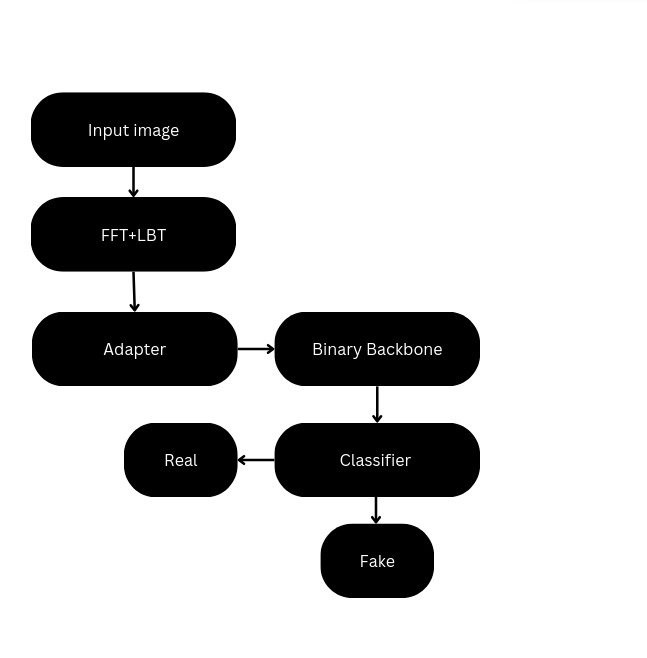}
  \caption{Workflow task 1}
  \label{fig:example}
\end{figure}

\subsubsection*{Model Selection}
Based on 2.1 and 2.2 sections we initially began working with EfficientNet and Faster Than Lies. 
\subsubsection*{\textbf{Training}}
We had the dataset divided in 2 classes \textbf{Real and Fake}.
We trained the models on CiFake to begin with. EfficientNetB1 displayed an accuracy of 94.1 {\%}. Faster than Lies displayed an accuracy of 95.7 {\%}. Post expansion of dataset we trained both the models parallely on following parameters: 
\begin{itemize}
    \item Initial Epochs: 20
    \item Train Images: 0.96 Million
    \item Test Images: 1.24 Million
    \item RandomHorizontal flip - 0.6
\item RandomVertical Flip - 0.6 prob
\item Random Erasing: 0.6 prob
\item RandomGrayscale - 0.6
\item RandomPerspective-0.6
\item Random rotation - 30 degress

\item Batch size 32 for Effiecient Net , 128 for Faster than Lies

    \item GPU used for EfficientNet: T4
    \item GPU used for Faster Than Lies: L4
\end{itemize}
We began with 20 epochs as the initial number which we constantly observed for overfitting. The EfficientNet model began overfitting at around 9 epochs and we did not let it go much beyond that so stopped training at 10 epochs.  The Faster than lies model showed overfitting at 4th epoch so we stopped training at the end of 5th epoch. 

\subsubsection*{Pertubation Identification:}
Since the problem statement also included adversarial
perturbations introduced to prevent detection. 

To tackle this issue, we followed a structured process:
\begin{enumerate}
    \item First, we analyzed the perturbations in a set of test images, evaluating their intensity based on noise, compression artifacts, and blur.
    \item Next, we extended this analysis to real-world images, as well as perturbed real images, to identify perturbations that exceeded a predefined threshold.
    \item Based on this analysis, we introduced similar perturbations to our training dataset, simulating realistic conditions and challenges.
    \item Finally, we retrained our model using this augmented dataset. By exposing the model to perturbed images, we allowed it to adapt to these variations and improve its robustness to such disturbances.
\end{enumerate}

\subsubsection*{\textbf{Perturbation Detection and Analysis}}

To accurately detect and analyze different types of perturbations, we developed a comprehensive set of methods. This included several image processing techniques designed to identify specific artifacts and distortions:

\begin{itemize}
    \item \textbf{Noise Detection:} We used Total Variation (TV) denoising and mean squared error (MSE) to assess noise in the images.
    \item \textbf{Compression Artifacts:} These were simulated by resizing and upscaling images, with intensity measured through MSE.
    \item \textbf{Blur Detection:} We quantified blur using the Laplacian variance of the image.
    \item \textbf{Adversarial Perturbations:} These were detected by analyzing edge intensity, using the Canny edge detection algorithm.
    \item \textbf{Color Perturbations:} We analyzed shifts in the red, green, and blue channels by calculating their absolute differences.
    \item \textbf{Saturation and Contrast Changes:} We examined the variance in the saturation channel of the image after converting it to HSV color space.
    \item \textbf{Pixel Shuffling:} This was simulated by randomly permuting the pixels of the image and measuring the resulting distortion using MSE.
    \item \textbf{JPEG Artifacts:} These were introduced by compressing images at lower quality levels and comparing the decompressed images to the originals.
    \item \textbf{Resizing Artifacts:} We simulated resizing artifacts by reducing the image size and then resizing it back, measuring the distortion through MSE.
    \item \textbf{Edge Smoothing:} We detected smoothing by comparing the original image to one blurred using a Gaussian kernel.
    \item \textbf{Motion Blur:} This was simulated by applying a motion blur-like effect to the image, again quantified with MSE.
    \item \textbf{Pattern Injection:} We detected repeating patterns in the frequency domain using Fourier transforms.
    \item \textbf{Brightness Adjustments:} We measured the mean pixel intensity to identify any significant changes in brightness.
\end{itemize}
\begin{figure}[htbp]
  \centering
  \includegraphics[scale=0.3]{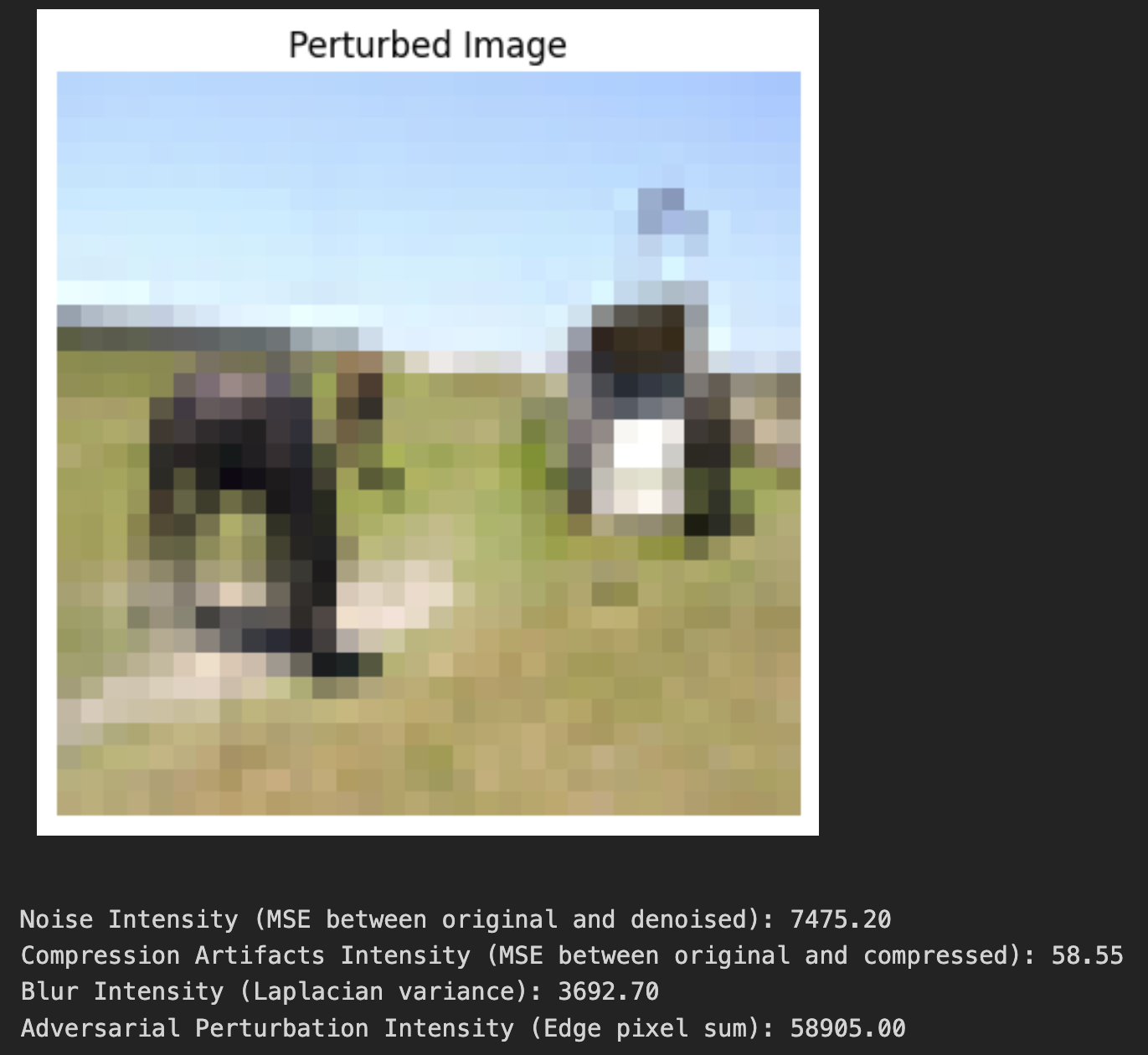}
  \caption{Pertubation Analysis}
  \label{fig:example}
\end{figure}
\begin{figure}[htbp]
  \centering
  \includegraphics[scale=0.3]{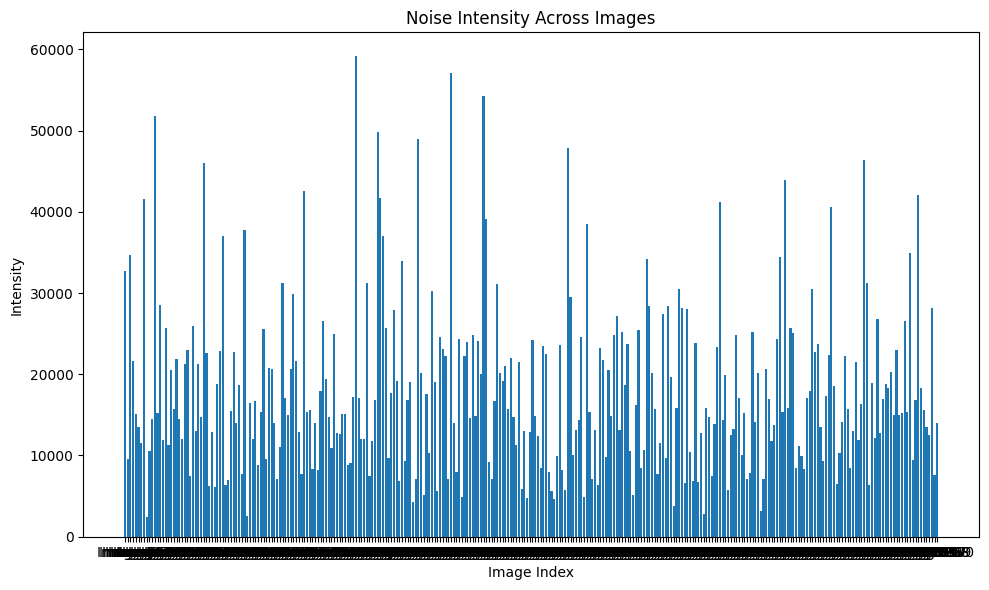}
  \caption{Pertubation Analysis}
  \label{fig:example}
\end{figure}
\begin{figure}[htbp]
  \centering
  \includegraphics[scale=0.3]{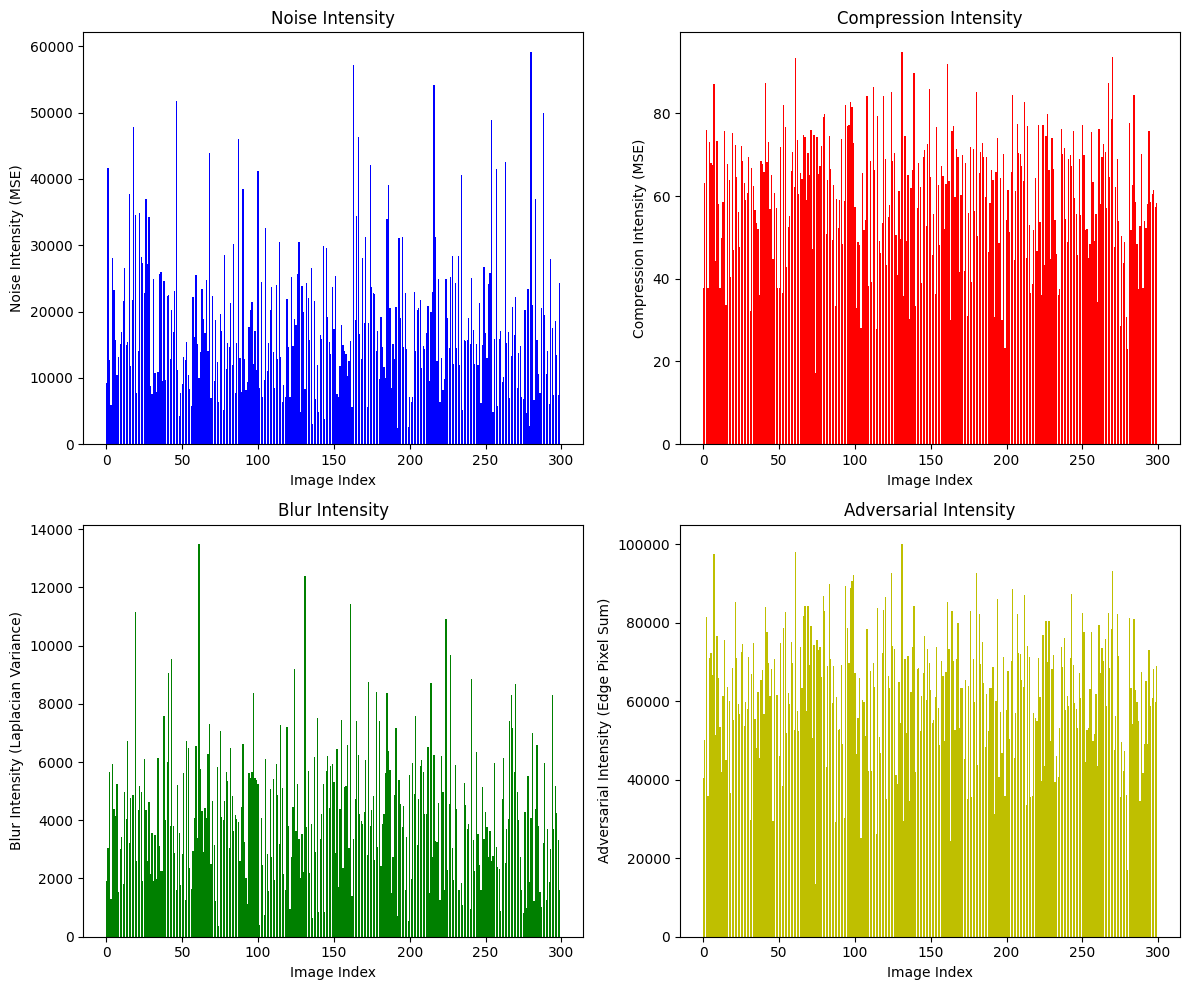}
  \caption{Pertubation Analysis}
  \label{fig:example}
\end{figure}

\subsubsection*{\textbf{Augmenting the Dataset and Retraining}}
After identifying and analyzing the perturbations, we augmented the CiFake training dataset with the following synthetic pertubations: gaussian noise, salt and pepper noise, motion blur, pixelate, random hue saturation, random erase, adversarial noise(using mobilenetv2),quantization artifacts, mask based corruption. 
This allowed the model to see more realistic perturbations, including noise, compression artifacts, and adversarial effects, during the training process.

Once the training dataset was enriched with these perturbed images, we retrained the model. This retraining helped the model become more robust by teaching it to recognize and adapt to various types of perturbations. As a result, the model's performance improved significantly, allowing it to handle adversarial conditions more effectively. 

\subsection*{Task-2:}
\begin{figure}[htbp!]
  \centering
  \includegraphics[scale=0.3]{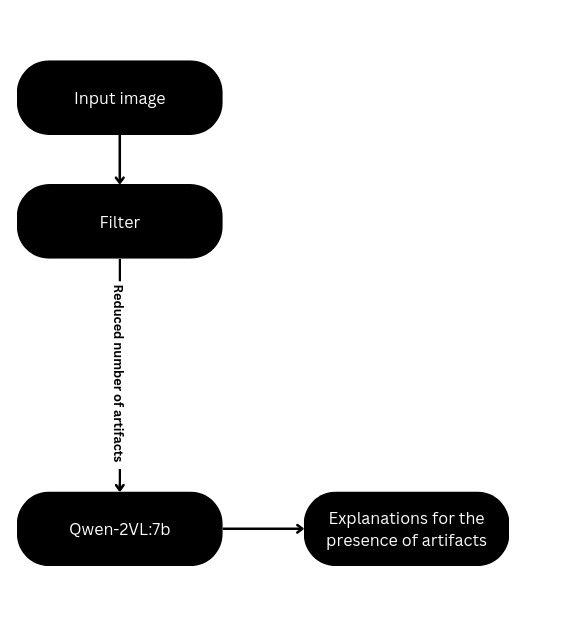}
  \caption{Workflow for task2}
\end{figure}

\subsubsection*{Model Selection:} Based on findings and experiments from Section 2.6 and Table 1, we decided to go ahead with Qwen2 VL 7B since it worked better for low resolution images and provided better explanation.  
\subsubsection*{Localisation: }
We trained an Autoencoder on Real images so that it learns how to reconstruct real images. We then reconstructed all images in our dataset , found pixelwise loss on each image. The real part had less loss since it could be reconstructed properly whereas the parts with higher loss could not be reconstructed properly and were plotted on attention map. The areas of higher loss were highlighted and considered the localisation of the artifact regions. These localised images gave better visual understanding to the VLMs as well as the humans making it more understandable.
\begin{figure}[htbp!]
  \centering
  \includegraphics[scale=0.3]{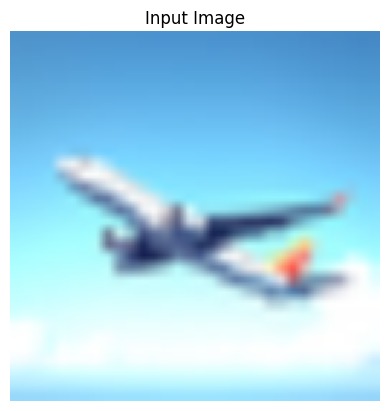}
  \caption{}
\end{figure}
\begin{figure}[htbp!]
  \centering
  \includegraphics[scale=0.3]{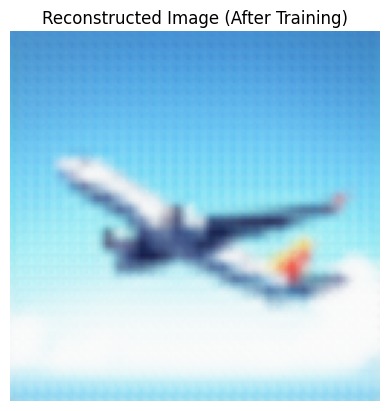}
  \caption{}
\end{figure}
\begin{figure}[htbp!]
  \centering
  \includegraphics[scale=0.3]{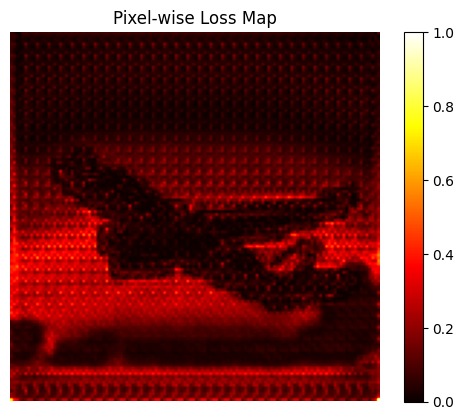}
  \caption{}
\end{figure}
\begin{figure}[htbp!]
  \centering
  \includegraphics[scale=0.3]{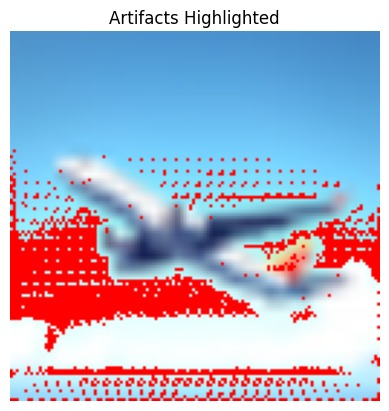}
  \caption{}
\end{figure}

\subsubsection*{Implementation and Results}

To detect and analyze the perturbations, we used a variety of image processing libraries, including OpenCV, NumPy, and SciPy. The results were visualized using matplotlib, which allowed us to track and compare perturbation intensities across different images. 

The retrained model showed substantial improvements, demonstrating a stronger ability to classify images even when they were distorted by various types of perturbations.  illustrates how perturbation intensities varied across different test images, showcasing the model's enhanced robustness.

\subsubsection*{\textbf{Artifact Filtering:}}
Out of the total 70 artifacts , we grouped them into 8 categories namely

\subsubsection*{Artifact Groups and Descriptions}

\subsubsection*{Geometric and Structural Anomalies}
\begin{itemize}
    \item Inconsistent object boundaries
    \item Discontinuous surfaces
    \item Non-manifold geometries in rigid structures
    \item Floating or disconnected components
    \item Asymmetric features in naturally symmetric objects
    \item Misaligned bilateral elements in animal faces
    \item Irregular proportions in mechanical components
    \item Impossible mechanical connections
    \item Inconsistent scale of mechanical parts
    \item Physically impossible structural elements
    \item Incorrect wheel geometry
    \item Implausible aerodynamic structures
    \item Misaligned body panels
    \item Impossible mechanical joints
    \item Anatomically impossible joint configurations
    \item Unnatural pose artifacts
    \item Biological asymmetry errors
    \item Excessive sharpness in certain image regions
    \item Unnaturally glossy surfaces
\end{itemize}

\subsubsection*{Texture and Surface Issues}
\begin{itemize}
    \item Texture bleeding between adjacent regions
    \item Texture repetition patterns
    \item Over-smoothing of natural textures
    \item Artificial noise patterns in uniform surfaces
    \item Metallic surface artifacts
    \item Artificial enhancement artifacts
    \item Regular grid-like artifacts in textures
    \item Repeated element patterns
    \item Synthetic material appearance
    \item Artificial smoothness
\end{itemize}

\subsubsection*{Lighting and Reflection Problems}
\begin{itemize}
    \item Unrealistic specular highlights
    \item Inconsistent material properties
    \item Multiple light source conflicts
    \item Missing ambient occlusion
    \item Incorrect reflection mapping
    \item Inconsistent shadow directions
    \item Glow or light bleed around object boundaries
    \item Incorrect Skin Tones
    \item Unnatural Lighting Gradients
    \item Dramatic lighting that defies natural physics
    \item Multiple inconsistent shadow sources
\end{itemize}

\subsubsection*{Anatomical and Biological Anomalies}
\begin{itemize}
    \item Dental anomalies in mammals
    \item Anatomically incorrect paw structures
    \item Improper fur direction flows
    \item Unrealistic eye reflections
    \item Misshapen ears or appendages
    \item Anatomically impossible joint configurations
    \item Impossible foreshortening in animal bodies
    \item Exaggerated characteristic features
\end{itemize}

\subsubsection*{Perspective and Spatial Distortions}
\begin{itemize}
    \item Incorrect perspective rendering
    \item Scale inconsistencies within single objects
    \item Spatial relationship errors
    \item Depth perception anomalies
    \item Fake depth of field
    \item Resolution inconsistencies within regions
    \item Artificial depth of field in object presentation
    \item Impossible mechanical joints
\end{itemize}

\subsubsection*{Image Quality Issues}
\begin{itemize}
    \item Over-sharpening artifacts
    \item Aliasing along high-contrast edges
    \item Blurred boundaries in fine details
    \item Jagged edges in curved structures
    \item Random noise patterns in detailed areas
    \item Loss of fine detail in complex structures
    \item Systematic color distribution anomalies
    \item Color coherence breaks
    \item Unnatural color transitions
    \item Frequency domain signatures
\end{itemize}

\subsubsection*{Visual Artifacts from Synthetic Image Generation}
\begin{itemize}
    \item Ghosting effects: Semi-transparent duplicates of elements
    \item Cinematization Effects
    \item Movie-poster like composition of ordinary scenes
    \item Unnatural pose artifacts
\end{itemize}

\subsubsection*{Occlusion and Object Cut-off Issues}
\begin{itemize}
    \item Abruptly cut off objects
    \item Inconsistent object boundaries
\end{itemize}

We passed all the fake detected images through a CLIP Encoder ViT-B32 which did basic visual interpretation of the images and depicted which category can the artifacts belong to with a certain confidence score. We then shortlisted the 3 highest scoring categories and parsed it to the VLM (Qwen2 7B).

\subsubsection*{Processing the images}
Qwen2 7B now looked for specifically these artifacts in the images. If detected , it did further visual analysis and created a textual interpretation of the artifact.

\section*{Results}
\subsection*{Inference Time}
\textbf{L40s} was the GPU we used. The following were the inference times:\begin{itemize}

\item Qwen2 VL 7B: 5.189s per image
     
\end{itemize}
On a 8 Core CPU:
\begin{itemize}
    
\item Image Localisation(Includes Reconstruction , Attention Mapping and Highlighting : 1s
\item Faster than Lies: 175ms 
\item EfficientNetB0: 136 ms
\item EfficientNetB0: 151 ms

\end{itemize}

\subsection*{Accuracy}:
\begin{itemize}
    \item Efficient net b0:  81 Percent (two class)
    \item Efficient net b3 74.53 Percent (4 class)
    \item Faster Than Lies without pertubations: 94 Percent
    \item faster than Lies with pertubations: 96.5 Percent
\end{itemize}
\subsubsection*{Qualitative Examples}
\begin{figure}[htbp]
  \centering
  \includegraphics[scale=0.5]{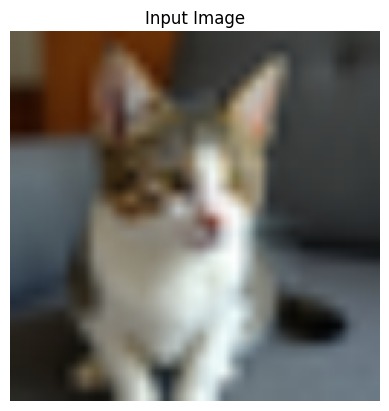}
  \caption{Input Image.}
  \label{fig:example}
\end{figure}

\textbf{Detected: Fake}
Top 3 detected artifact categories:
\begin{itemize}
    \item Occlusion and object cutoff Issues
    \item Visual artifacts from synthetic image generation
    \item Occlusion and object cutoff Issues
\end{itemize}
\textbf{Index:} 18

\textbf{Explanation:}
\begin{itemize}
    \item \textbf{Over-sharpening artifacts:} The image appears to have excessive sharpening, causing the edges and details to be overly pronounced.
    \item \textbf{Random noise patterns in detailed areas:} Noticeable noise patterns are present, particularly in the fur of the cat, disrupting the natural texture.
    \item \textbf{Aliasing along high-contrast edges:} There is aliasing along the edges of the cat's fur, especially around the ears and eyes, resulting in a jagged appearance.
    \item \textbf{Frequency domain signatures:} The image exhibits high-frequency noise patterns, which are visible in the fur texture.
    \item \textbf{Color coherence breaks:} Abrupt color transitions are observed, especially around the edges of the cat's fur, indicating a lack of smooth color transitions.
\end{itemize}

\section*{Discussion}
\subsection*{Limitations of current solution}
\begin{itemize}
    \item The explanations are dependednt on VLM 
    \item Might be more AI image sources whoch we did not incorporate 
    \item We might be missing on some pertubations
    \item The accuracy of the second solution's explanations cannot be tested on the current dataset provided to us.
    \item Our solution might not work on images smaller than this.
    \item Localisation has not fully been acheived , it can be implemented in a  better way by even deeper analysis.
    \item Speed vs. Accuracy Trade-off: There is an inherent trade-off be-
tween the speed of the classifier and its accuracy. Optimizing the
classifier for faster performance negatively impacts its accuracy,
and vice versa. This creates a limiting factor in finding the right
balance between speed and accuracy.

\end{itemize}
\subsection*{Observations regarding the data}
The test data had 185 fake images and 115 Real Images , all heavily pertubated. 
Common Artifacts:
\begin{itemize}
    \item \textbf{Inconsistent shadow sources}: Shadows do not align with a single light source, a common artifact in AI-generated images.
    \item \textbf{Dramatic lighting}: Overly dramatic lighting that defies natural physics.
    \item \textbf{Incorrect reflection mapping}: Reflections do not match the lighting conditions.
    \item \textbf{Inconsistent shadow directions}: Shadows point in different directions, indicating inconsistent lighting.
\end{itemize}

\begin{itemize}
    \item \textbf{Inconsistent material properties}: Mismatch in textures, e.g., rough vs. smooth fur or mixed materials (wood, metal).
    \item \textbf{Artificial smoothness}: Over-smoothing that removes fine details.
    \item \textbf{Texture repetition patterns}: Repetitive patterns in the texture due to artificial generation.
\end{itemize}

\begin{itemize}
    \item \textbf{Incorrect perspective rendering}: Distorted proportions and unrealistic object alignment.
    \item \textbf{Scale inconsistencies}: Objects appear disproportionately large or small.
\end{itemize}
\begin{itemize}
    \item \textbf{Loss of fine detail}: Compression or resolution issues causing loss of intricate details.
    \item \textbf{Random noise patterns}: Noise visible in detailed areas of the image.
    \item \textbf{Over-sharpening artifacts}: Excessive sharpening, creating exaggerated edges.
    \item \textbf{Artificial depth of field}: Unnatural depth of field effects.
\end{itemize}

\subsection*{Implementation Difficulties}

Developing an efficient model for classifying and explaining 32x32 resolution images presented several challenges, as most existing research focuses on higher-resolution images. Key difficulties included:

\begin{itemize}
      
    \item \textbf{Low-Resolution Image Classification:} Classifying low-resolution images was difficult due to limited visual information. Existing models designed for higher-resolution images were ineffective, requiring the development of new architectures and techniques.
    \item \textbf{Lightweight and Locally Deployable Model:} The model had to be computationally efficient while maintaining accuracy, making it suitable for mobile or embedded devices with limited resources.
    \item \textbf{Explainability of Classifications:} Low-resolution images limited the effectiveness of Vision-Language Models (VLMs) for generating explainable results. Significant effort went into prompt engineering and fine-tuning to improve explainability.
    \item \textbf{Handling Pertubations} Handling pertubations was a major task since they were intended to fool the classification model. A lot of research went into understanding the noises that might affect the detection process and in synthetically creating them.
    \item \textbf{Dataset Limitations:} Scarcity of annotated datasets for artifact detection hindered model fine-tuning. We had to curate and augment data to overcome this challenge.
    \item \textbf{Selection of Vision-Language Models:} Choosing the right VLM for low-resolution classification and artifact explanation involved extensive evaluation and testing.
    \item \textbf{Computational Constraints:} The lack of free GPU resources slowed training and fine-tuning, limiting our ability to quickly test different models or architectures.
\end{itemize}
\subsection*{Potential
Improvements}
Based on the challenges and opportunities identified during the implementation process, the following potential improvements could enhance the solution:

\begin{itemize}
    \item \textbf{Dynamic Resolution Image Support:} \\
    While the model performs well for 32x32 resolution images, expanding the solution to handle higher-resolution images (e.g., 64x64 or 128x128) could improve classification accuracy. This would be particularly beneficial for cases where more detailed information is available.

    \item \textbf{Enhanced Model Efficiency:} \\
    Although the model is lightweight, exploring additional optimizations such as quantization or pruning could further reduce the model’s size and inference time. These optimizations would be especially beneficial for deployment on edge devices with limited resources, enabling faster and more efficient operations on CPUs or low-power GPUs.

    \item \textbf{Improved Explainability:} \\
    Despite significant efforts in prompt engineering, enhancing the explainability of classifications could be achieved by integrating attention mechanisms or saliency maps. These techniques would allow for a better understanding of the important features used by the model in its decision-making process, making the model’s predictions more interpretable, especially in real-world applications.

    \item \textbf{Data Augmentation and Expansion:} \\
    The scarcity of annotated datasets for artifact detection presents a limitation. Expanding the dataset by collecting more labeled data or leveraging semi-supervised learning techniques could substantially improve the model’s robustness. Additionally, introducing synthetic data generation methods (e.g., using GANs) could augment the training dataset and improve performance on previously unseen data.

    \item \textbf{Model Evaluation with Diverse VLMs:} \\
    While Qwen2 VL 7B was selected, further experimentation with a broader range of Vision-Language Models (VLMs) could uncover models better suited for this task. Models like CLIP or VisualBERT may offer better generalization capabilities, especially for low-resolution images, and should be evaluated for potential improvements.

    \item \textbf{Improved Handling of Artifact Detection:} \\
    Given the importance of artifact detection, integrating additional techniques like anomaly detection or contrastive learning could enhance the model’s ability to distinguish between artifacts and genuine features. This would improve the model's robustness to real-world variations and perturbations in the data.

    \item \textbf{Real-Time Inference Improvements:} \\
    For applications requiring low-latency responses (e.g., edge devices or mobile applications), improving the inference pipeline for real-time applications could be beneficial. Techniques such as model distillation or hardware-specific optimizations could be explored to reduce inference time further.

    \item \textbf{Cross-Domain Generalization:} \\
    To improve robustness across various domains (e.g., different types of images, artifacts, or tasks), the model could be fine-tuned on a more diverse set of data from different sources. This would help ensure that the model generalizes well to new or unseen contexts, expanding its applicability.
\end{itemize}

These improvements would help increase the model's accuracy, efficiency, explainability, and robustness, making it more versatile and applicable to a broader range of real-world use cases.
\subsection*{Broader Applications}
\begin{itemize}
    \item \textbf{Medical Imaging:} Enhance diagnosis with low-resolution images from X-rays or MRIs, detecting anomalies in resource-limited environments.
    \item \textbf{Industrial Inspection:} Detect defects and wear in machinery or infrastructure using low-resolution inspection imagery.
    \item \textbf{Art Preservation:} Classify and analyze low-resolution images of historical artifacts for conservation and restoration.
    \item \textbf{Social Media Moderation:} Enhance content moderation by classifying low-resolution images for inappropriate content.
    \item \textbf{Disaster Recovery:} Aid recovery efforts by analyzing low-resolution imagery to identify damaged areas.
    \item \textbf{Forensic Analysis:} Support criminal investigations with the analysis of low-resolution security footage and public records.
\end{itemize}
\section*{Conclusion and Future Work}
We presented an explainable system for detecting AI-generated imagery that integrates lightweight convolutional networks with Vision–Language Models for human-interpretable explanations. The model’s robustness to adversarial perturbations and its deployability on local devices highlight its practical potential. Future work includes extending the approach to variable-resolution datasets, enhancing localization precision through diffusion-based reconstruction, and evaluating cross-domain generalization across social, forensic, and industrial datasets.

\cite{ricker2024aeroblade,wang2023dire,Lanzino_2024_CVPR,cao2024synartifact,guo2018countering,xu2024llavacot,qwen,liu2023improvedllava,liu2023llava,liu2024llavanext,xu2024fakeshield,tan2019efficientnet,dosovitskiy2021image}
\bibliographystyle{ACM-Reference-Format}

\end{document}